\documentclass[runningheads]{llncs}

\usepackage{helvet}
\usepackage{courier}
\usepackage{graphicx}
\usepackage{listings}
\usepackage{xcolor}
\usepackage{tikz}
\usepackage{booktabs}
\usepackage{array}
\usepackage{longtable}

\usepackage{graphicx}
\begin{document}
\title{Building Scaffolding Dialogue Data with LLM-Simulated Novices}

\titlerunning{SimInstruct}

%
%
\author{
Si Chen \and
Izzy Molnar \and
Ting Hua \and
Peiyu Li \and
Le Huy Khiem \and
G.~Alex Ambrose \and
Jim Lang \and
Ronald Metoyer \and
Nitesh V.~Chawla
}

\institute{
University of Notre Dame, USA\\
\email{
schen34@nd.edu,
imolnar@nd.edu,
thua@nd.edu,
pli9@nd.edu,
kle3@nd.edu,
gambrose@nd.edu,
jlang4@nd.edu,
rmetoyer@nd.edu,
nchawla@nd.edu
}
}

\maketitle              
\author{
Si Chen \and
Izzy Molnar \and
Ting Hua \and
Peiyu Li \and
Le Huy Khiem \and
G.~Alex Ambrose \and
Jim Lang \and
Ronald Metoyer \and
Nitesh V.~Chawla
}

\institute{
University of Notre Dame, USA\\
\email{
schen34@nd.edu,
imolnar@nd.edu,
thua@nd.edu,
pli9@nd.edu,
kle3@nd.edu,
gambrose@nd.edu,
jlang4@nd.edu,
rmetoyer@nd.edu,
nchawla@nd.edu
}
}

\begin{abstract}

High-quality, multi-turn instructional dialogues between novices and experts are essential for developing AI systems that support teaching, learning, and decision-making. These dialogues often involve \textit{scaffolding}—the process by which an expert supports a novice’s thinking through questions, feedback, and step-by-step guidance. However, such data are scarce due to privacy concerns in recording and the vulnerability inherent in help-seeking.
We present \textsc{SimInstruct}, a scalable, expert-in-the-loop tool for collecting scaffolding dialogues. Using teaching development coaching as an example domain, \textsc{SimInstruct} simulates novice instructors via LLMs, varying their teaching challenges and LLM's persona traits, while human experts provide multi-turn feedback, reasoning, and instructional support. This design enables the creation of realistic, pedagogically rich dialogues without requiring real novice participants.
Our results reveal that persona traits—such as extroversion and introversion—meaningfully influence how experts engage. Compared to real mentoring recordings, \textsc{SimInstruct} dialogues demonstrate comparable pedagogical relevance and cognitive depth. Experts also reported the process as engaging and reflective, improving both data quality and their own professional insight.
We further fine-tuned a LLaMA model to be an expert model using the augmented dataset, which outperformed GPT4o in instructional quality. Our analysis highlights GPT4o’s limitations in weak reflective questioning, overuse of generic praise, a condescending tone, and a tendency to overwhelm novices with excessive suggestions.
\end{abstract}
\section{Introduction}
Scaffolded multi-turn dialogues, where support is gradually adjusted based on learner need, form the basis of effective learning in many domains \cite{wood1976role,van2010scaffolding}. These interactions foster metacognition, adapt to individual needs, and encourage reflective reasoning rather than direct answer delivery. Scaffolding plays a central role not only in formal education like mathematics instruction~\cite{collins1989cognitive}, but also in informal and professional development contexts—including clinical supervision, counseling, peer mentoring, and coaching—where nuanced feedback and situated reasoning are essential~\cite{tran2025generative}.

As interest in AI for education grows, so does demand for language models that provide multi-turn scaffolding rather than straightforward fact delivery. OpenAI's recently introduced Study Mode is one such example: a tutoring-like interface that uses Socratic prompts and scaffolded feedback across multiple turns to foster active reflection and deeper learner engagement~\footnote{https://openai.com/index/chatgpt-study-mode/}. Other promising systems—such as CLASS and KG-RAG—demonstrate scaffolded, stepwise QA and measurable gains in education~\cite{CLASS2023,dong2025build}.

Despite the demand, authentic scaffolding dialogues remain elusive at scale. Because these interactions are often private, ethically sensitive, and highly context-dependent—especially in education—they are rarely documented or generalizable~\cite{lee2022coauthor}. Moreover, traditional data collection relies on post-hoc annotation of small recordings or static LLM pipelines that omit expert intent and situated pedagogical reasoning~\cite{lee2022coauthor,geiger2021garbage}. As a result, coaching-style instructional dialogue datasets remain scarce.

In this work, we introduce \textsc{SimInstruct}, a responsible and scalable tool for generating expert–novice scaffolding dialogues using arge language model(LLM)-simulated personas. Human experts engage in multi-turn conversations with a wide range of LLM-simulated novices, offering guidance, reflection, and instructional reasoning. This expert-in-the-loop design produces pedagogically rich, contextually grounded dialogue data while avoiding the ethical and logistical challenges of recruiting real novices. We evaluate the quality of 123 dialogues from 18 experts by comparing them to real coaching interactions in teacher development programs and find that \textsc{SimInstruct} achieves comparable quality. 

\textbf{Our contributions are as follows:}
(1) We propose \textsc{SimInstruct}, a novel expert-in-the-loop and responsible data collection tool that leverages LLM-simulated novices to elicit rich, multi-turn instructional dialogues grounded in scaffolding theory.
(2) We evaluate the collected dialogues through both qualitative feedback and quantitative analysis. By comparing our data with real expert–novice recordings and fine-tuned LLMs on the augmented dataset, we demonstrate improved instructional quality and pedagogical relevance.
(3) We distill practical lessons for involving human experts in high-quality dataset creation and offer actionable insights for building domain-specific AI.

Among the many domains and subjects where scaffolding is critical, we focus on \textit{teacher development} as our case study. In both preservice training and ongoing professional learning, expert–novice coaching helps educators interpret classroom events, make instructional decisions, and develop their pedagogical identity~\cite{hennessy2019teacher,wallin2015impact}. Recent research shows that existing tools often fall short in addressing the real-world complexities of teaching, with educators expressing a preference for AI systems that offer more contextualized and reflective support~\cite{chen2025bridging}. We answer the research question--\textit{How effectively does the use of LLM-simulated novices support the generation of high-quality scaffolding dialogues?}

\section{Background and Related Work}

\subsection{Human–AI Collaboration in Dataset Collection}
Human–AI collaboration has been widely applied in dataset creation, especially in annotation workflows where humans curate or label AI-generated content, as seen in projects like Soylent and Garbage-in, Garbage-out~\cite{bernstein2010soylent,geiger2022garbage}. With the rise of large language models (LLMs), newer pipelines generate synthetic data across domains including healthcare~\cite{li2023chatdoctor}, math tutoring~\cite{tang2023mathdial}, and instruction tuning~\cite{wang2022selfinstruct}.

Despite these advancements, many pipelines treat LLMs as autonomous generators, with humans only reviewing or filtering content after generation~\cite{lee2022coauthor}. Approaches such as GLAN~\cite{li2024glan} and MIDI-Tuning~\cite{wang2024midi} improve consistency through structured prompting and multi-round tuning but still fall short in integrating domain expertise or pedagogical objectives.

In contrast, recent HCI research demonstrates how domain experts can play a more central role in AI-supported workflows. Yang et al.\cite{yang2021data} emphasize the co-construction of datasets with AI to enhance interpretability and contextual alignment. Collaborative tools like DocuViz \cite{kim2014docuviz} visualize writing dynamics to scaffold expert coordination, while Mechanical Novel~\cite{kim2017mechanicalnovel} shows how experts and non-experts can co-author content through iterative reflection and revision.

\subsection{Scaffolding in Teaching and Learning Systems}

Effective teaching involves more than delivering correct answers. Scaffolding strategies—such as metacognitive prompting, formative feedback, and adaptive coaching—help learners construct knowledge and build autonomy~\cite{van2012scaffolding,chi2014icap}. Yet many conversational AI systems prioritize factual accuracy or task completion, reinforced through training methods like reinforcement learning from human feedback~\cite{ouyang2022training}.

Such systems risk “answer-leaking,” which can bypass productive struggle and hinder learning~\cite{koedinger2012data}. Although datasets like MathDial incorporate step-by-step reasoning~\cite{tang2023mathdial}, they fall short of modeling the coaching-based interactions essential in teacher development. Our work builds on these insights by simulating expert–novice scaffolding dialogues that reflect authentic instructional goals. Recent benchmarks like ClarifyBench assess multi-turn reasoning but lack pedagogical context~\cite{lin2023clarifybench}. 

\subsection{Expert–Novice Dialogues in Professional Development}
Expert–novice interactions are central to professional learning in fields such as education, counseling, and clinical supervision~\cite{schon1983reflective,eraut2000non}. In \textit{teacher development}, coaching conversations help novices interpret classroom events, navigate uncertainty, and cultivate their pedagogical identity. However, capturing these dialogues at scale remains difficult due to privacy concerns and logistical constraints~\cite{perell2021designing}.

Beyond supporting novice growth, such interactions benefit experts as well by promoting reflection and reinforcing instructional judgment. This dual impact aligns with the learning through teaching paradigm, which suggests that teaching others enhances the teacher’s own understanding by requiring knowledge reorganization, articulation of tacit reasoning, and anticipation of learner needs~\cite{bargh1986learning,chi2014icap}. These benefits have been documented across teacher preparation, medical supervision, and design-based learning~\cite{eraut2000non,schon1983reflective}. Recent work has begun to explore how LLM-simulated learners can support expert tasks. For example, Generative Students~\cite{lu2024generativestudents} introduces simulated student profiles to aid item evaluation in assessment design, illustrating how synthetic learners can enhance expert workflows. 

\section{Method}  
We adopted a human-centered design approach to develop \textit{SimInstruct}, a web-based tool for collecting multi-turn instructional dialogues between human experts and LLM-simulated novice instructors. From January to June 2025, we conducted weekly design sessions with two senior teacher development experts to iteratively refine the tool’s interface, persona generation, and study protocol. In July 2025, human experts used \textit{SimInstruct} asynchronously, generating dialogues.  We evaluated the tool through two lenses:  (1) content analysis of the collected dialogues and their suitability for fine-tuning AI models (AI perspective); and  (2) expert feedback on usability and perceived realism (HCI perspective).  All procedures were approved by the Institutional Review Board.

\begin{figure*}[t]
\includegraphics[width=8cm]{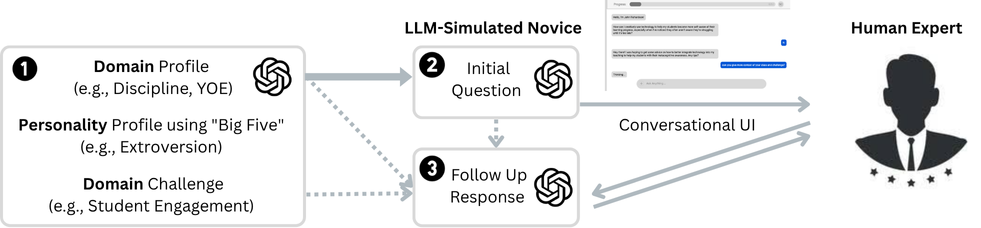}
\centering
\caption{\textsc{SimInstruct} includes a conversational UI that allows human experts to engage in multi-turn dialogues with LLM-simulated novices across a wide range of persona profiles at their own pace.}
\label{fig:process}
\end{figure*}

\section{\textsc{SimInstruct} Tool Design \& Development}

\subsection{Design Goals}

\textbf{G1: LLM-Simulated Novices to Mitigate Privacy Risks}
We initially aimed to collect traditional expert–novice teaching dialogues via audio recording, but privacy concerns, scheduling difficulties, and worries that being recorded might inhibit authentic and often vulnerable help-seeking limited us to being able to collect just four sessions (around 200 minutes) over two months. To address ethical and logistical barriers, \textsc{SimInstruct} uses LLMs to simulate novice instructors to talk with human experts, enabling the scalable collection of realistic dialogues without relying on sensitive or identifiable data.  These four real dialogues were later used to check the quality of the dialogues collected with our tool, for example in Fig \ref{fig:his}.

\textbf{G2: Conversational Data Collection to Capture Expert Reasoning Naturally} We chose a conversation-based approach to data collection because it better captures the dynamics of real instructional interactions and allows experts to express their reasoning more naturally, engagingly and flexible than through static annotations or scoring tasks.

\textbf{G3: Persona-Based for Diverse Contexts and Questions} We created diverse, pedagogically meaningful personas to simulate novices with varied subject areas, experience levels, and instructional challenges, enabling the generation of dialogue data across a broad range of teaching contexts. We also varied personality traits, such as introversion and extroversion, to further diversify conversational behaviors and content. The goal was to make the dialogues resemble authentic novice–expert interactions.


\subsection{\textsc{SimInstruct} Data Collection Tool Details} 
\begin{figure}[t]
\includegraphics[width=12cm]{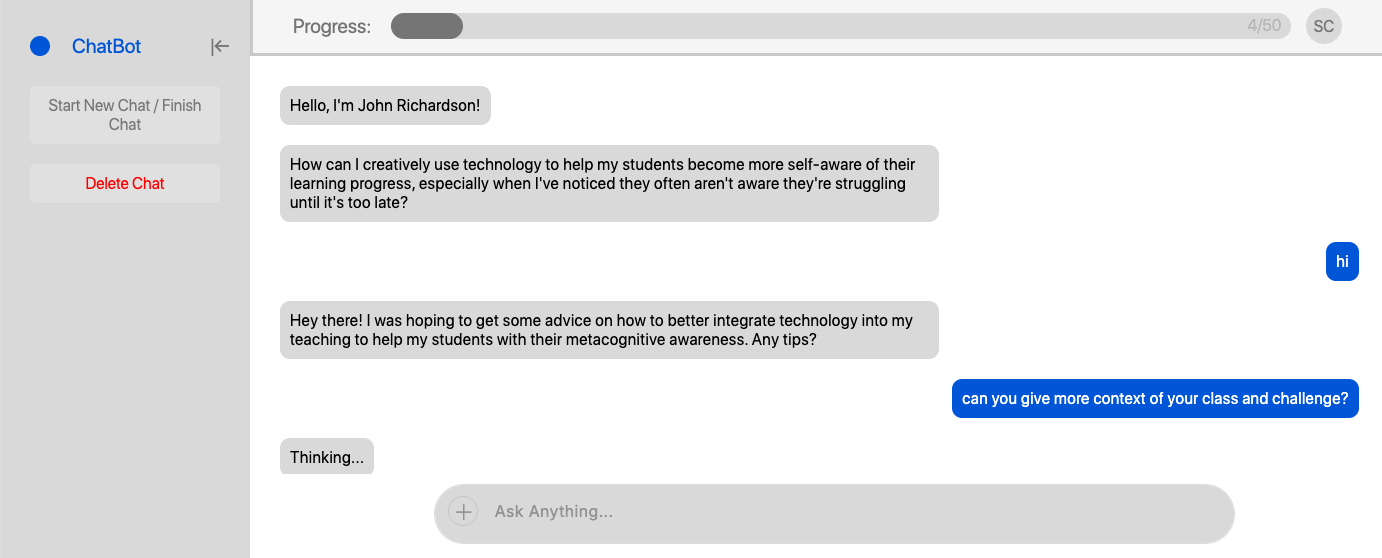}
\centering
\caption{\textsc{SimInstruct} UIUX}
\label{fig:UI}
\end{figure}

\textsc{SimInstruct} is designed to simulate realistic novices, enabling human teaching experts to engage in conversations that resemble typical coaching interactions—offering an alternative to real instructors, thereby addressing \textit{G1}. To support \textit{G2}, we developed a web-based conversational user interface (UI) (Full version in Fig.\ref{fig:UI}) where experts interact with an LLM-simulated novice through multiple dialogue turns. As illustrated in Fig.~\ref{fig:process}, the simulated novices are assigned diverse personas with traits relevant to the teacher coaching domain, addressing \textit{G3}. Prompts can be found in Appendix A.

Below, we provide more details on the three key components of LLM-Simulated Novice. \tikz[baseline=(char.base)]{
\node[shape=circle, draw=black, fill=black, text=white, inner sep=1pt] (char) {1};
}
\textbf{Persona Profile Generator}
Each simulated novice is assigned a persona comprising nine randomly selected attributes for \textit{Domain Profile}: first name, last name, classroom context, teaching experience, discipline, course level, semester context, teaching style, and conversation style. Then for \textit{Personality Profile}: Teaching and conversation styles are defined using four of the Big Five personality traits—openness, conscientiousness, extroversion, and agreeableness~\cite{mccrae1987validation}—which are randomly assigned. Neuroticism was excluded as it was considered irrelevant to our task. A \textit{Domain Challenge} was then randomly selected from a 40 item list created by human expert. To ensure internal consistency, a verification function using GPT-4 checks each generated profile for logical coherence (e.g., avoiding mismatches like a law professor assigned to a laboratory classroom). \tikz[baseline=(char.base)]{
\node[shape=circle, draw=black, fill=black, text=white, inner sep=1pt] (char) {2};
}
\textbf{Initial Question}
Once the persona is generated, GPT-4 is used to produce an initial question, which is presented to the human expert. The expert must respond based solely on the information provided, eliciting further details through follow-up questions. This interaction design reflects a realistic coaching scenario and supports \textit{G2}. \tikz[baseline=(char.base)]{
\node[shape=circle, draw=black, fill=black, text=white, inner sep=1pt] (char) {3};
}
\textbf{Follow-up Responses}
For subsequent turns, GPT-4-turbo-preview generates responses aligned with the persona profile and initial question to behavior coherence. The expert continues the dialogue with the simulated novice until they determine the conversation is complete and a viable teaching strategy has been identified—or that the novice appears ready to implement and satisfied with experts' suggestion.

When engaging with the tool, experts are presented with a welcome message from the simulated novice, which includes the novice's first and last name along with the initial question. Experts interact with a unique set of novice profiles, with the option to delete any conversations they find unrealistic.  To refine the generation process, the two human expert was asked to review and evaluate 30 randomly selected novice profiles which included the persona and initial question. The expert’s feedback was incorporated to enhance realism and coherence of the generated profiles.

\subsection{Data Collection Study with Human Experts}
Eighteen human experts based in the US, all with extensive coaching experience in higher education, participated in the study. They were recruited through email and word-of-mouth to ensure their expertise and experiences. All expert had advanced degree. Human experts completed their dialogues asynchronously over a two-week period. Compensation was based on the number of completed dialogues, with per-dialogue pay calculated according to the estimated time taken (5 to 20 minutes) and the participant’s hourly rate of $50$ USD. Throughout the study, participants were encouraged to communicate with the research team to ask questions, raise concerns, and share positive experiences. At the end of the study, all participants were invited to complete a brief post-hoc survey to provide feedback on the tool, the quality of the dialogues, and their experience.

\subsection{Description of Collected Data}





\subsubsection{Overview} 
\begin{figure*}[t]
\includegraphics[width=12cm]{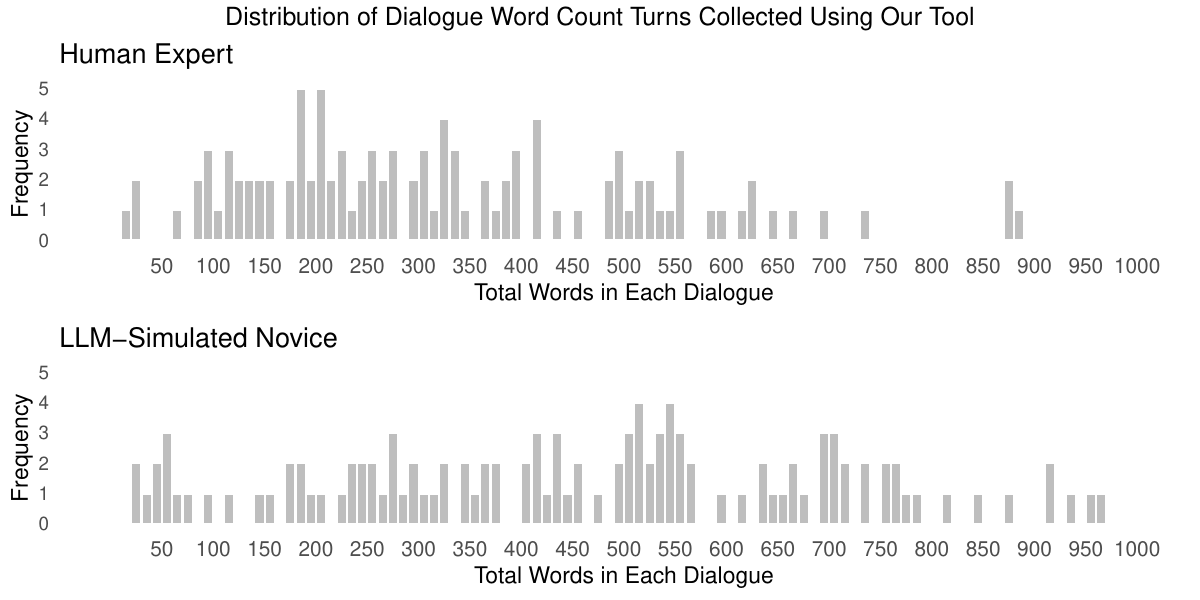}
\centering
\caption{Our tool collected wide word count range in human expert and LLM-simulated novice dialogues.}
\label{fig:tworole}
\end{figure*}

Across 123 dialogues, the LLM-simulated novice produced a total of 65,004 words, and the human expert contributed 38,444 words, with 1,848 total dialogue turns. On average per dialogue, the LLM contributed 528.49 words (SD = 378.40), the human expert contributed 312.55 words (SD = 242.41), and there were 15.02 turns (SD = 7.77). The distribution of the dialogues turns can be seen in Fig. \ref{fig:his}. A ``turn'' refers to a single speaker's utterance. More data can be seen in Fig. \ref{fig:tworole}. The majority dialogues collected through our tool followed a three-part scaffolding structure: Problem Identification, where the novice raised a specific teaching challenge; Reason Exploration, where the expert guided reflection on underlying causes; and Strategy Development, where practical solutions were discussed. A sample dialogue is found in Appendix B.

\subsubsection{Findings} We further analyzed dialogue turns by examining novice's different persona input across different aspects. 
Human expert word count was analyzed in relation to four LLM-simulated novice personality traits (extraversion, agreeableness, conscientiousness, openness) while accounting for repeated contributions from the same expert. Records with missing values or expert\_word\_count $\leq$ 100 were removed, and trait values were recoded as Low vs. High. For each trait, a linear mixed-effects model was fit with trait level as a fixed effect and a random intercept for expert\_Id:
\texttt{expert\_word\_count \textasciitilde{} trait\_level + (1 | expert\_Id)}. High–Low contrasts with 95\% confidence intervals were extracted. Multiple-comparison correction was not applied because each trait was tested as a separate, theory-motivated hypothesis. 

\begin{figure}[t]
\centering
\begin{minipage}{0.48\linewidth}
  \centering
  \includegraphics[width=\linewidth]{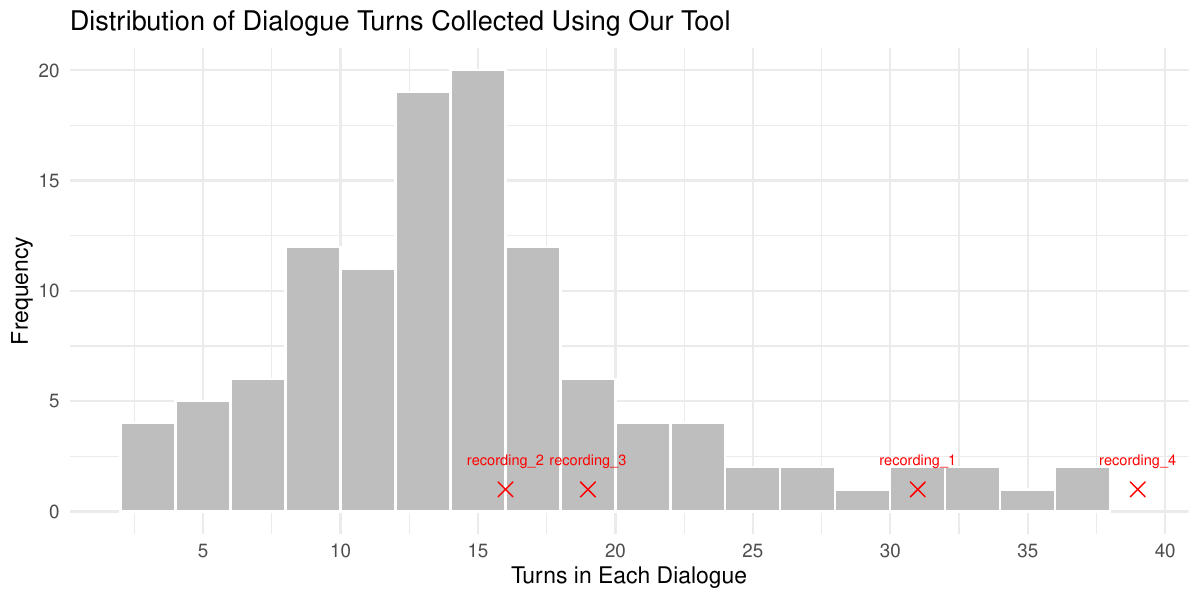}
  \caption{Our tool collected 123 dialogues with diverse turn counts, with a median of approximately 15 turns. The red X represents data from four real recorded sessions.}
  \label{fig:his}
\end{minipage}
\hfill
\begin{minipage}{0.48\linewidth}
  \centering
  \includegraphics[width=\linewidth]{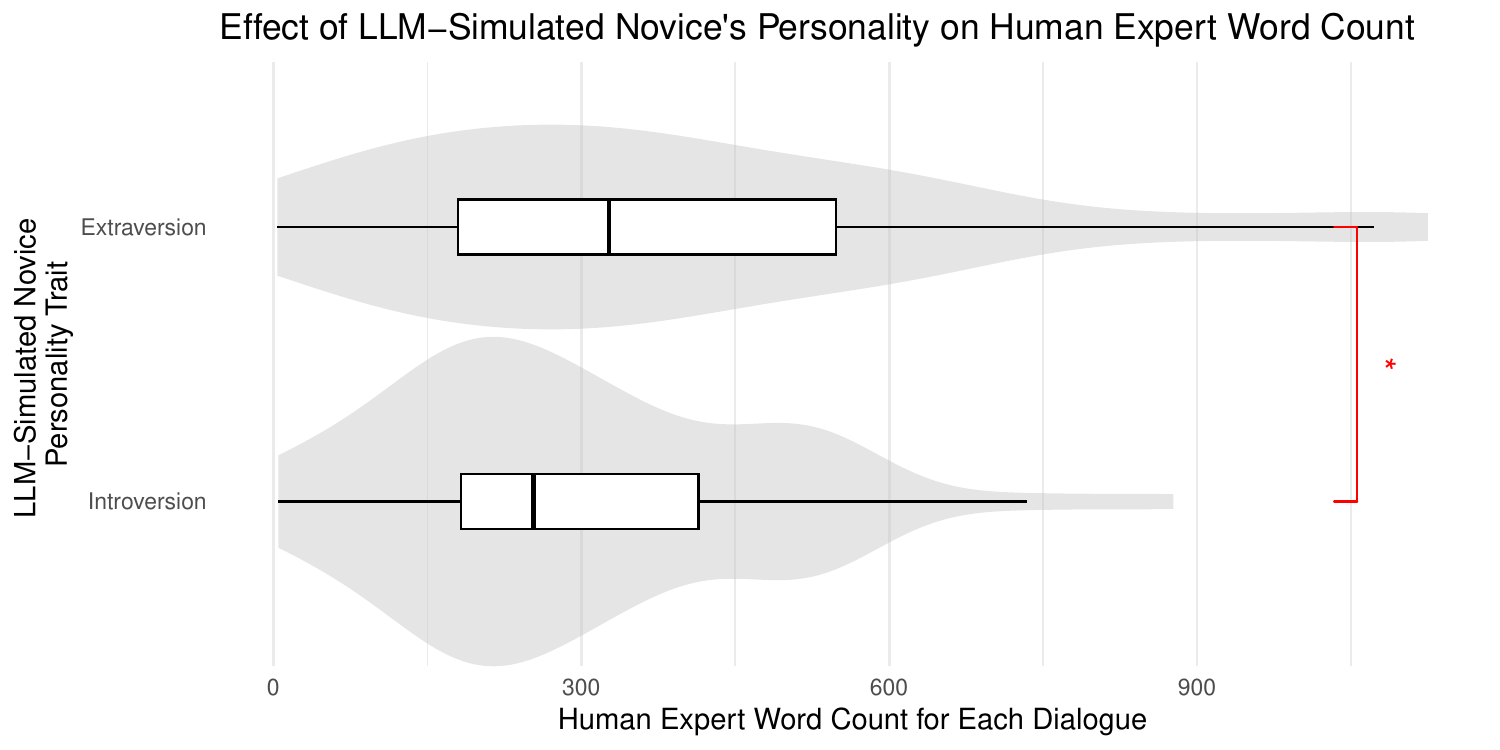}
  \caption{Extroverted personas of LLM-simulated novices led human experts to engage in more conversation compared to introverted personas, demonstrating the impact of LLM persona design on data collection.}
  \label{fig:persona}
\end{minipage}
\end{figure}

\begin{figure*}[t]
\includegraphics[width=12cm]{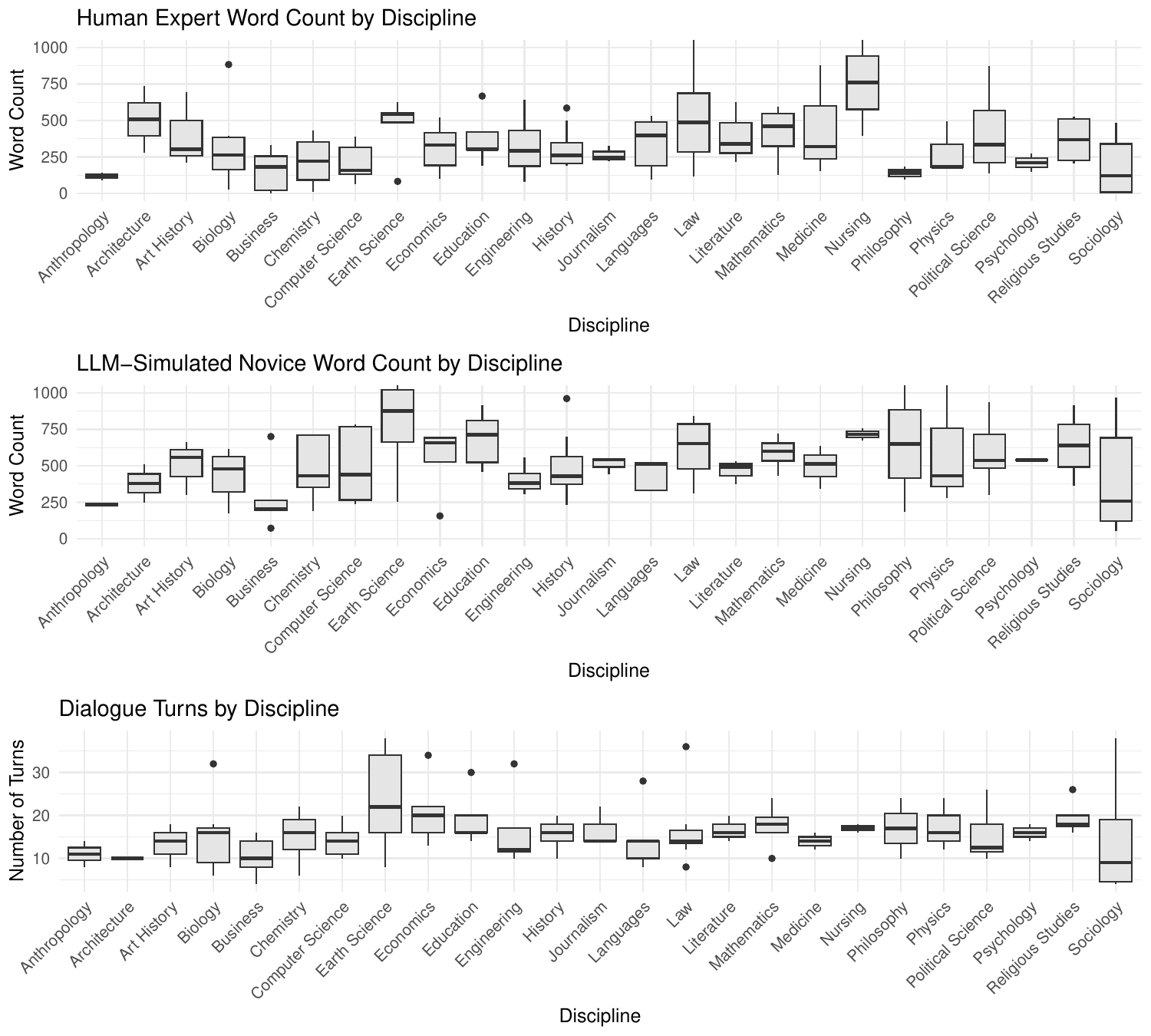}
\centering
\caption{ur tool collected wide word count range of discipline.}
\label{fig:displine}
\end{figure*}

Across 100 dialogues,  extraversion showed a positive association with expert word count ($\Delta=86.94$, $p=0.030$) (Fig. \ref{fig:persona}), whereas agreeableness ($\Delta=-35.60$, $p=0.374$), conscientiousness ($\Delta=5.82$, $p=0.893$), and openness ($\Delta=9.04$, $p=0.823$) showed no reliable effects. Discipline (Fig. \ref{fig:displine}): Fields like Earth Science and Nursing showed the longer average dialogues, while Anthropology, Business, and Sociology had shorter exchanges. This variation may reflect differing conversational needs or how experts' own background and cannot be generalized.

\section{Evaluation of \textsc{SimInstruct}}

\subsection{SimInstruct Collected Data vs. Real Dialogues}
We compared four dialogues collected using our tool with four real, face-to-face coaching sessions totaling of 200 minutes. While this comparison is limited in scope due to the small sample size, it offers preliminary insights into both the promise and limitations of using LLMs to simulate novice behavior. As shown in Fig. \ref{fig:his}, approximately half of the collected dialogues fall within the turn range observed in real dialogues, while the other half exhibit fewer turns. We did not compare word counts, as face-to-face oral dialogues often include filler words and self-correction language, that are not typically present in text communication.

\subsubsection{Human Annotators’ Rating of Collected Data}
To evaluate the quality of the dialogues, human expert raters assessed each dialogue using a rubric based on human-developed evaluation criteria. Four evaluation criteria were developed collaboratively by team members with experience in pedagogy and educational research and were used to judge how well the simulator replicated realistic and meaningful novice questions. These criteria capture overall pedagogical relevance, cognitive depth of the simulated question, instructional contextualization, and coverage of pedagogical concerns. Two researchers independently rated those dialogue data with with a $\kappa$ of 0.69. A list of these evaluation criteria is found in Appendix C.

\subsubsection{Findings}
\begin{figure}[t]
\centering
\begin{minipage}{0.48\linewidth}
  \centering
  \includegraphics[width=\linewidth]{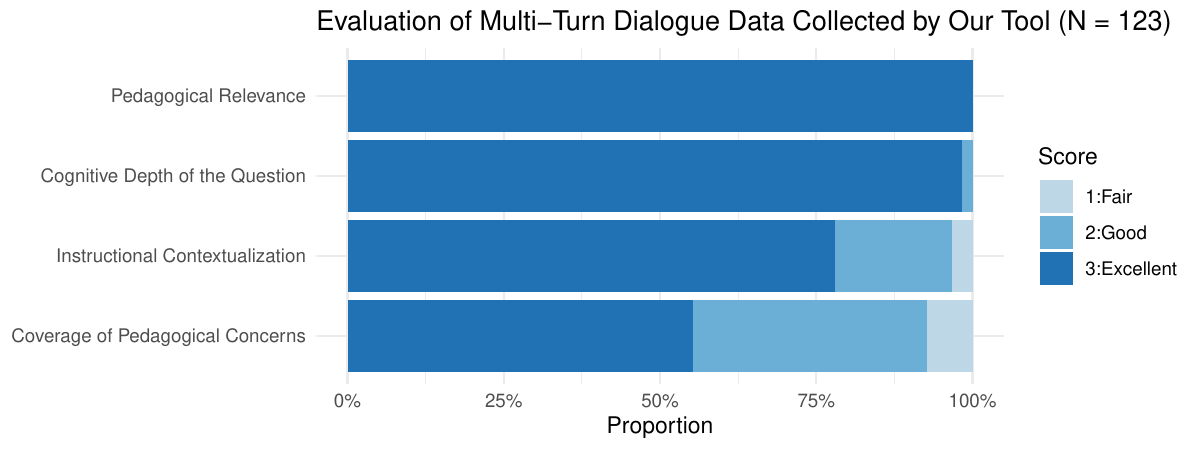}
  \caption{Human evaluation shows good to excellent dialogue quality in the collected novice data. LLM-simulated novice challenges were sometimes straightforward, overly direct and sometimes lacked contextual information in the questions asked.}
  \label{fig:stackedplot}
\end{minipage}
\hfill
\begin{minipage}{0.48\linewidth}
  \centering
  \includegraphics[width=\linewidth]{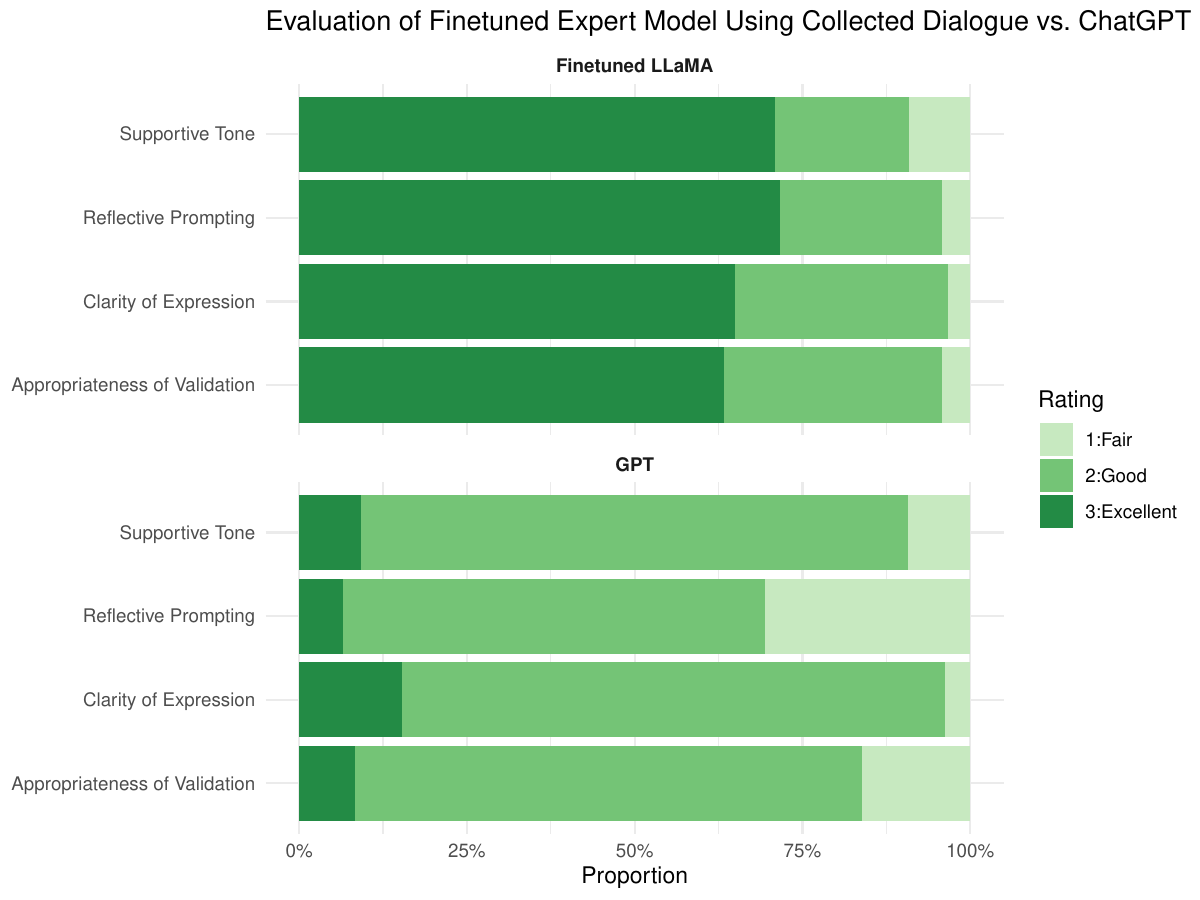}
  \caption{Human ratings show our finetuned LLaMA outperforms GPT4o, particularly in reflective prompting—ability to ask thoughtful questions.}
  \label{fig:tworatertwomodel}
\end{minipage}
\end{figure}

\begin{figure}[t]
\centering
\includegraphics[width=0.6\linewidth]{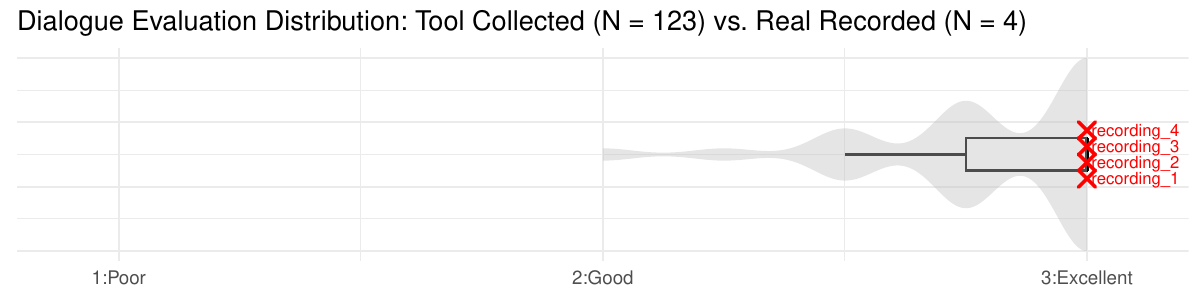}
\caption{Compared to real recorded dialogues rated excellent (3 out of 3 in all four criteria), dialogues collected by our tool averaged 2.80 (SD = 0.25), slightly lower. The red X represents data from four real recorded sessions.}
\label{fig:average}
\end{figure}

We found that the structural scaffolding in both types of dialogues was similar. In particular, both data included the three-step scaffolding structure: identifying a problem, reasoning about it, and suggesting a strategy. The average scores across the four criteria were consistently lower for the \textsc{SimInstruct} collected data compared to the face-to-face recorded data. As shown in Fig. \ref{fig:average}, recorded dialogues consistently received scores of 3 across all criteria, while the \textsc{SimInstruct} collected data had slightly lower averages. A closer look at the score distribution of the \textsc{SimInstruct} data revealed more specific patterns as seen in Fig. \ref{fig:stackedplot}. While the ratings show that the tool-generated dialogues as containing high-quality pedagogical features, the LLM-simulated novices in these dialogues sometimes appeared overly straightforward and lacked contextual grounding of own challenges. This suggests that the simulated novice responses may not yet fully replicate the complexity of human novices.

\subsection{Collected Data to Finetune Expert Model} While our tool collected a set of high quality expert-novice dialogues, the dataset still remains relatively small and insufficient for fine-tuning of language models. To address this limitation, we decided to augment the data using synthetic generation strategies.

\subsubsection{Data Augmentation}
Starting with a seed dataset of 123 collected dialogues, we use GPT‑4o mini to generate over 1,000 synthesized dialogues within a consistent domain. In each iteration, we randomly sample three real dialogues from the seed set as examples and prompt the language model to generate a new multi-turn dialogue in the same format and with a similar tone and style, but beginning with a different initial question. This process leverages the model’s in-context learning capabilities. After synthesis, we filter out any dialogues that do not match the required format, resulting in a final set of 1,415 dialogues.

\subsubsection{Finetune Expert Model using Augmented Data}
We fine-tuned the \texttt{Llama-2-7b-chat-hf} model using the full augmented dataset described previously, consisting of 1,415 multi-turn scaffolding dialogues. After splitting these into individual turns, we obtained a total of 9,271 training examples. Fine-tuning was performed on a single NVIDIA A100 GPU for 435 steps, taking approximately 1.5 hours. We used a learning rate of $2 \times 10^{-5}$, weight decay of 0.01, warmup ratio of 0.05, and cosine learning rate scheduling. Optimization was done using the AdamW optimizer from PyTorch.

\subsubsection{Human Annotators' Rating of Expert Model} 
\begin{figure*}[t]
\includegraphics[width=16cm]{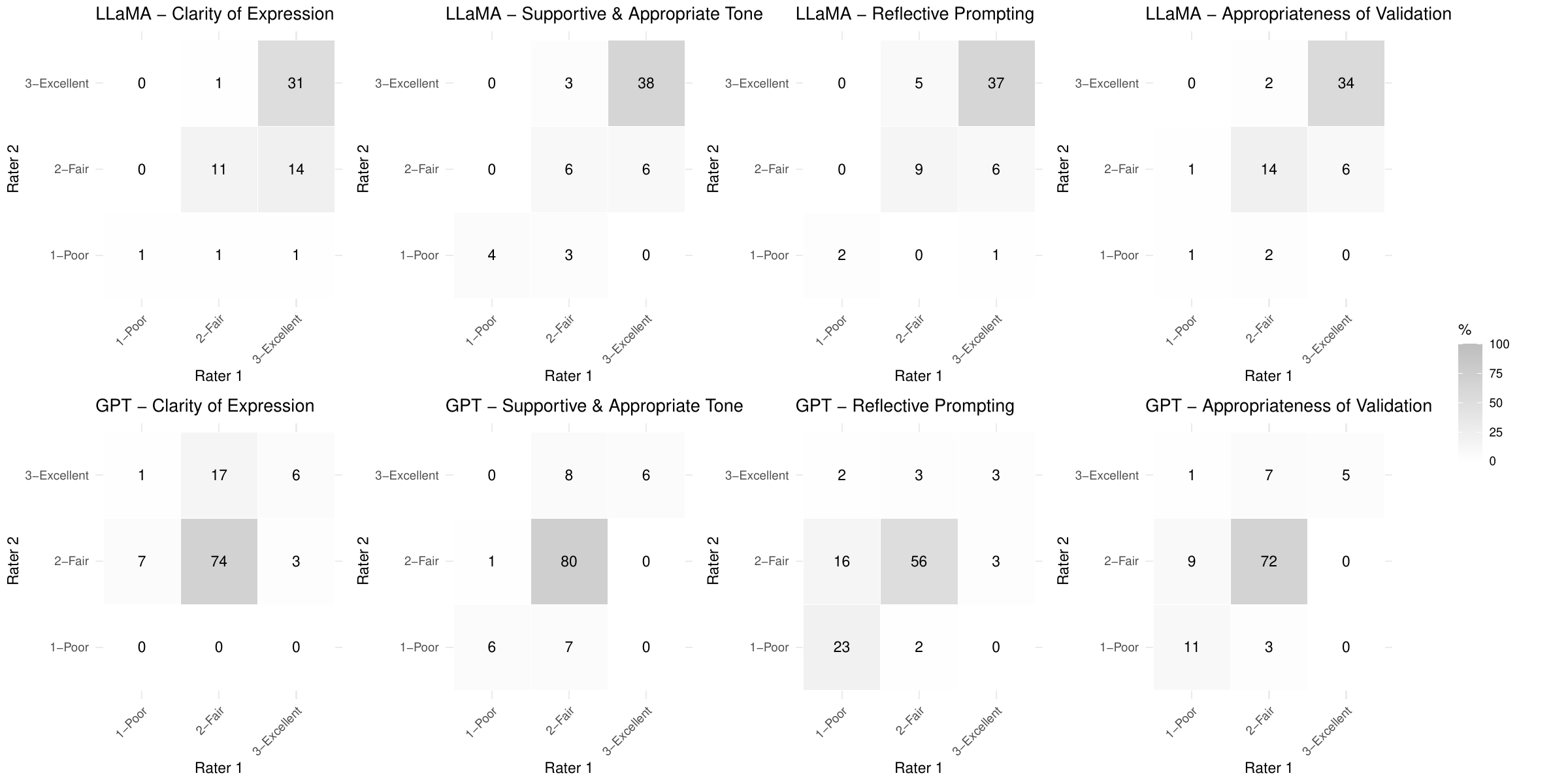}
\centering
\caption{Rater 1 and Rater 2 showed higher overall agreement when evaluating LLaMA responses ($\kappa$ = 0.65) than GPT responses ($\kappa$ = 0.53), indicating greater consistency in judgments across all dimensions.}
\label{fig:matrix}
\end{figure*}

We conducted a human evaluation of 220 AI-generated instructional dialogues, with the source model of each dialogue kept hidden and shuffled. Two human annotators independently rated the expert responses for each dialogue. We used a rubric developed by our research team with extensive expertise in teaching and AI. The rubric focuses on how effectively expert responses coach novice instructors, using four criteria on a 1-3 scale (full definitions provided in Appendix D): Clarity of Expression, Supportive \& Appropriate Tone, Reflective Prompting, and Appropriateness of Validation. To assess inter-rater reliability across these dimensions, we computed quadratically weighted Cohen’s $\kappa$ on the pooled ratings. LLaMA achieved a $\kappa$ of 0.65, indicating substantial agreement, while GPT reached a moderate agreement level with a $\kappa$ of 0.53. See Fig. \ref{fig:matrix} for detailed agreement matrices. 


\subsubsection{Findings} As show in Fig \ref{fig:tworatertwomodel}, LLaMA outperformed GPT4o across all four evaluation dimensions. The most pronounced difference was in reflective prompting, where LLaMA received a substantially higher mean score (M = 2.67, SD = 0.55) than GPT4o (M = 1.76, SD = 0.56), indicating that LLaMA was significantly more effective at encouraging novices to reflect or reason. In terms of clarity of expression and supportive tone, LLaMA again showed stronger performance, achieving mean scores of 2.62 on both, compared to GPT4o’s 2.12 and 2.00, respectively. Finally, for appropriateness of validation, LLaMA scored 2.59 (SD = 0.57) while GPT4o scored 1.92 (SD = 0.49), suggesting LLaMA provided more balanced and contextually appropriate validation without overusing generic praise. These results indicate that LLaMA’s responses were consistently clearer, more supportive, more reflective, and more appropriately validating than those of GPT4o.

\subsection{Human Expert's Experience of Using the Tool} 
Expert participants reported a positive experience using the tool, as reflected in both informal interactions and a post-study survey. The lightweight, asynchronous interface was described as easy to navigate, and the conversation-based format aligned with their coaching practices. Survey responses indicated that most participants found the simulated scenarios realistic and the task intellectually engaging. Several noted that the process prompted reflection on their own coaching styles and approaches to feedback. Experts appreciated the flexibility to skip scenarios, and many expressed interest in using similar tools for training or professional development.

    



 \subsubsection{Encouraging Expert Participation Through Framing and Flexible Task Design} Framing the task as a meaningful contribution to the future of AI in education was key to recruiting and sustaining expert engagement. We aligned the task with familiar elements of their daily routines and avoided presenting it as a generic data labeling exercise. Instead, we emphasized that their participation would directly inform how AI tools support real-world instruction—positioning their expertise as central to AI design.
To accommodate diverse work styles and schedules, we decided to do flexible compensation options, including task-based (per-dialogue) payment. This flexibility signaled respect for expert labor and contributed to higher engagement and completion rates.

\subsubsection{Engaging with AI as Reflective Learning Opportunity}

Creating novel and intellectually engaging data collection tools was essential for meaningful expert participation. Rather than assigning routine annotation tasks, we invited experts to engage in authentic, multi-turn instructional dialogues that mirrored real coaching practice. This positioned them as co-designers contributing pedagogical insight, rather than passive data labelers. Participants had full autonomy to skip, revise, or decline prompts they found unproductive, reinforcing respect for their professional judgment. This opt-out flexibility supported sustained engagement and high-quality contributions.

Many human experts described the task as more reflective and rewarding than traditional data collection efforts. The interactive format encouraged deeper professional reflection and engagement. These findings suggest that expert-centered design—grounded in authenticity, flexibility, and intellectual value—is essential for responsible and effective data collection in education and beyond.

\begin{quote}
    \textit{``...I would prefer my feedback to be used not only to make a case for the possibilities (positive outcomes) of AI in teaching and learning but to also inform consequences (negative implications) to consider as educators.''} -- Human Expert in Post-Hoc Survey
\end{quote}


Across workshops, experts flagged criteria that appeared objective but lacked instructional nuance. For instance, a relevance item asking whether the expert directly answered the novice’s question was unanimously rejected, as effective teaching often involves redirecting or prompting deeper reflection. All groups independently converged on similar revisions, underscoring consistent pedagogical reasoning. This process highlights the value of LLMs not as final evaluators, but as prompts for expert reflection that strengthened the evaluation framework.

\subsection{Promises and Challenges of LLM for Teaching and Learning}
Our results show that dialogue turns and word counts span a wide range. Notably, Fig. \ref{fig:persona} illustrates how extroverted and introverted LLM-simulated novice personas influenced expert behavior—even when the dialogues were similar in length and number of turns. This highlights the importance of persona design in creating realistic data and suggests that personality traits can shape how human experts engage with LLMs and the kinds of responses they provide.

Another challenge, illustrated in Fig. \ref{fig:stackedplot} and reinforced by expert feedback, is that the LLM-simulated novice sometimes posed overly straightforward instructional challenges—ones that could be resolved with a single suggestion. Experts observed that human novices are more likely to say “no” or push back on suggestions, whereas the LLM consistently agreed, even when they are prompted to decline when appropriate. As a result, coaching the LLM felt easier but less authentic than coaching real novices for human experts. These gaps underscore the need for future LLM systems to better simulate disagreement and dialogic complexity, especially since LLMs are typically optimized to be agreeable responders.



Third, Fig. \ref{fig:tworatertwomodel} shows that GPT was less capable of asking meaningful, reflective questions—an essential aspect of effective teaching that supports metacognitive thinking in learners. This raises important concerns about the extent to which AI, especially current GPT, can serve as a good teacher. In addition to limited reflective prompting, we observed other problematic behaviors such as overusing generic praise, adopting a condescending tone, and overwhelming learners with an excessive number of suggestions. These patterns may hinder the learning process.

Last, human-human face-to-face dialogues also often included more turns as shown in Fig. \ref{fig:his} and shorter utterances like ``keep going,'' which were absent in any dialogues collected in our tool. These brief responses, often tied to non-verbal cues, signaled idea-seeking and created natural pauses. It opens a new discussion on how human-human data can or cannot can be used to train LLM doing similar things as human.

\textbf{Limitations}
For this study, we selected teacher coaching as the example domain, rather than subjects like mathematics. While math-based tasks often emphasize factual correctness, teacher coaching focuses on reasoning an reflection that are more broadly transferable across instructional contexts. 
That said, transferring \textsc{SimInstruct}  to other domains still requires domain-specific adaptation. Persona profiles and challenges must be defined in collaboration with domain experts to ensure realism. Additionally, the domain-specific language—including how challenges are described, what counts as effective feedback, and how success is framed—would need to be re-authored to reflect norms in fields such as law, medicine, or engineering. We intentionally excluded gender and age from persona prompts to avoid introducing potential biases in simulated behavior—an area that merits further research.

\section{Conclusion} \textsc{SimInstruct} offers a responsible and scalable approach for collecting expert–novice scaffolding dialogues using LLM-simulated personas. Our findings demonstrate that expert engagement, thoughtful persona design, refined evaluation criteria, and expert-in-the-loop dialogue generation can yield pedagogically meaningful data while addressing common privacy and ethical concerns. Fine-tuning LLaMA with this augmented dataset improved its instructional quality over GPT, particularly in reflective prompting—its ability to ask thoughtful questions. However, challenges remain: LLM behaviors such as excessive agreement, user-pleasing tendencies, and limited contextual nuance hinder realistic simulation. 

%

\bibliographystyle{splncs04}
\bibliography{references}

\clearpage

\onecolumn
\section{Appendix}

\subsection{Appendix A: Prompt Used in Data Collection Tool}

\begin{lstlisting}[frame=single, caption=\textbf{Initial Question:}, breaklines=true]
"""Based on this teaching profile:
\{profile\_text\}

Generate a single, clear question about \{challenge\} in the motivation of \{category\}.
The challenge is: \{description\}
Sample question for inspiration: \{sample\_question\}

Requirements:
1. Use plain, direct language - write like you're talking to a trusted colleague, do not use formal/stilted language
2. Include emotional honesty - don't shy away from expressing genuine teaching concerns
3. Present a clear teaching dilemma - show the tension between different teaching approaches or goals
4. Avoid jargon and very domain-specific terms; use language that any educator could understand
5. The question should be specific and actionable
6. It should end with a question mark
7. It should be directly related to the challenge and profile
8. It should be a single question, not multiple questions
9. The question should be in first person (I, my, etc.)
10. Feel free to explain prior attempts to fix the challenge (i.e. \"I've tried... and it didn't work\")
11. Avoid self-deprecating language, describe your challenge in a way that a normal instructor would
12. If you talk about previous experiences, make sure to describe these experiences

Example of good tone:
I'm struggling to balance my need for structure with students' need for flexibility. How can I maintain clear expectations while being more responsive to their needs?

Return only the question, nothing else."""
\end{lstlisting}

\begin{lstlisting}[frame=single, caption=
\textbf{Follow Up Response:
}, breaklines=true]
"""You are an instructor in a coaching conversation with a teaching expert. Your goal is to learn and apply new strategies to your teaching.

Remember: This is a natural conversation between colleagues, not a formal presentation. Sounding human is more important than sounding perfect. You should sound like a human orally, not written language. 

When responding:
- Be concise and to the point
- Use simple language and avoid complex words
- Use short sentences and avoid long ones
- Use simple words and avoid complex words
- Use short sentences and avoid long ones
- Be authentic and let your personality shine through
- Share your real thoughts, uncertainties, and excitement
- Share specific details when relevant (duration, number of students, materials)
- If something doesn't feel right or you're unsure, ask
- Consider your time, resources, and students' needs
- Ask questions, but not every turn needs to be a question
- Focus on addressing the expert's latest message and your challenge/initial questions
- Ask follow up questions to the expert's suggestions so that you are able to implement them in your teaching
- Feel free to turn down the expert's suggestions if your don't think they are a good fit for your teaching style or if you don't have the time to implement them, it is normal to turn down 1-4 suggestions.
- Responses must not exceed 5 sentences

You have the following teaching profile:
\{conversation\_instructor\_profile\}"""

\end{lstlisting}
\clearpage

\clearpage

\subsection{Appendix B: Sample Instructional Conversations Collected Using Data Collection Tool}

\begin{quote}
    
          {
            "role": "assistant",
            "content": "How can I effectively engage my students in the decision-making process of the course content without compromising my structured teaching approach?",
          },

          {
            "role": "user",
            "content": "Hi Bob! What do you teach and what level are your students?",
          },

          {
            "role": "assistant",
            "content": "I teach advanced Biology, primarily to college students. It's an online course, so keeping them engaged and giving them some control while sticking to a structured outline is a bit of a challenge for me.",
          },

          {
            "role": "user",
            "content": "Have you tried surveying your students to learn more about their goals for the class?",
          },

          {
            "role": "assistant",
            "content": "Not yet, but that sounds like a good idea. I've been hesitant because I'm not sure how to balance their interests with the course's set curriculum. How detailed should these surveys be? Do you think a simple questionnaire at the beginning of the semester would do the trick, or should it be more ongoing?",
          },

          {
            "role": "user",
            "content": "I think the beginning of the course is a great opportunity for a detailed survey. Then, mid-semester could be another opportunity to do a follow-up survey that checks-in with students to see if needs have been met. ",
          },

          {
            "role": "assistant",
            "content": "That makes sense. Starting with a detailed survey could help me understand their interests and expectations early on. And a mid-semester check-in sounds like a great way to adjust if needed while still keeping things structured. I’ll need to figure out the right questions that allow me to gather useful information without promising to steer too far from the curriculum. Thanks for the suggestion!",
          },

          {
            "role": "user",
            "content": "Yes, you could also create assignments that offer students choice. For example, students could choose to give a presentation, write a research paper, conduct a podcast etc. This allows your students to have agency over their education and will help to keep them more engaged if they get a choice.",
          },

          {
            "role": "assistant",
            "content": "Offering choices in how they complete assignments is a brilliant idea. It aligns well with my goal to maintain structure while also giving students some control over their learning. I'll need to think about how each option can meet the learning objectives of the course. This way, regardless of the format they choose, the core content remains consistent. It's a good compromise. Thanks for the suggestion!",
          },

          {
            "role": "user",
            "content": "I'm happy to help! Along with choice, you can offer opportunities for students to work with a partner or small group to accomplish a task. This way you can maintain your structure but get students more engaged with their peers. You can allow them to have breakout rooms during class for planning and then it will be up to them to find ways to meet online outside of class. Will they work for your class structure?",
          },

          {
            "role": "assistant",
            "content": "Incorporating group work sounds like a good strategy to boost engagement and peer interaction, especially in an online setting. Using breakout rooms for initial planning is something I can do. However, given my preference for a highly structured environment, I might need to set clear guidelines and objectives for these group activities to ensure they stay on track. I'm a bit concerned about how they'll coordinate outside of class, but encouraging them to use common communication tools could help. I'll give it a shot and see how it goes. Thanks for the advice!",
          },

          {
            "role": "user",
            "content": "Yes, you should provide students with a list of tasks to accomplish within their breakout rooms. This will keep them focused. Students can make it work meeting outside of class time, whether that's through a shared Google Doc or other tool, or actually meeting online. Offer students a variety of methods to accomplish this; they'll make it work. ",
          },

          {
            "role": "assistant",
            "content": "That's reassuring to hear. Providing a clear list of tasks for the breakout rooms seems like a solid way to ensure productivity and focus. I like the idea of suggesting different methods for collaboration outside class, like Google Docs or online meetings, to accommodate various schedules and preferences. It sounds like with the right guidance and tools, students can effectively manage their group work. I'll start planning how to integrate this and ensure clear instructions are in place. Thanks for the encouragement and practical tips!",
          },

          {
            "role": "user",
            "content": "You're welcome! Is there anything else I can help you with today?",
          },

          {
            "role": "assistant",
            "content": "I think you've given me a lot to start with, especially on engaging students more actively in their learning process while keeping the structure I value. I'll work on implementing these strategies and see how they pan out. Thanks again for all the advice and support!",
          },

          {
            "role": "user",
            "content": "Have a great day!",
          },

          {
            "role": "assistant",
            "content": "You too! Thanks for all the help. Have a fantastic day!",
          }

\end{quote}

\clearpage

\clearpage

\subsection{Appendix C: Teacher Simulator Criterion For \textsc{SimInstruct}}
\label{AppendixE}

\begin{longtable}{>{\raggedright\arraybackslash}p{3.2cm} >{\raggedright\arraybackslash}p{4.2cm} >{\raggedright\arraybackslash}p{2.7cm} >{\raggedright\arraybackslash}p{2.7cm} >{\raggedright\arraybackslash}p{2.7cm}}
\toprule
\textbf{Criterion} & \textbf{Guiding Question} & \textbf{3} & \textbf{2} & \textbf{1} \\
\midrule
\textbf{Pedagogical Relevance} & Does the simulator ask questions that are appropriate for an instructor to ask when planning, reflecting on, or troubleshooting instruction? & Question aligns with common instructional challenges, planning, or feedback needed. & Question is tangentially related to instruction but lacks focus or depth. & Question is unrelated to pedagogical practice or classroom concerns. \\
\midrule
\textbf{Cognitive Depth of the Question} & Does the simulator’s question require the pedagogical expert to provide insight at an appropriate level of depth? & Asks for reasoning, strategy, tradeoffs, or adaptation to context. & Asks for clarification, examples, or basic explanation. & Asks for trivial, obvious, or yes/no responses. \\
\midrule
\textbf{Instructional Contextualization} & Is the question framed with enough instructional context (e.g., class type, learner profile, course goals) to allow the expert to respond meaningfully? & Question includes enough information (e.g., setting, learners, challenge) for a targeted response. & Context is vague or partial. & Lacks context; the expert must guess or infer too much. \\
\midrule
\textbf{Coverage of Pedagogical Concerns} & Over time, does the simulator generate questions across a range of instructional areas (e.g., assessment, student engagement, scaffolding, feedback)? & Addresses multiple aspects of teaching practice. & Covers a few areas but not comprehensive. & Focuses narrowly on one type of teaching concern. \\
\bottomrule
\end{longtable}

\subsection{Appendix D: Human Annotator Criterion for Expert Model}

\begin{longtable}{>{\raggedright\arraybackslash}p{3.2cm} >{\raggedright\arraybackslash}p{4.2cm} >{\raggedright\arraybackslash}p{2.7cm} >{\raggedright\arraybackslash}p{2.7cm} >{\raggedright\arraybackslash}p{2.7cm}}
\toprule
\textbf{Criterion} & \textbf{Guiding Question} & \textbf{3} & \textbf{2} & \textbf{1} \\
\midrule
\endfirsthead
\toprule
\textbf{Criterion} & \textbf{Guiding Question} & \textbf{3} & \textbf{2} & \textbf{1} \\
\midrule
\endhead

\textbf{Clarity of Expression} & Is the expert’s language clear, free of jargon, and concise enough to be quickly understood by instructors with varied teaching experience and English proficiency? 
& Clear and informative: Well-phrased, avoids jargon, easily understood. 
& Understandable: Some jargon or assumptions of prior knowledge; could be more concise. 
& Vague or confusing: Hard to follow, overly technical, or verbose. \\
\midrule

\textbf{Supportive \& Appropriate Tone} & Does the expert use an encouraging tone that feels appropriate for adult learners, without sounding overly directive or condescending? 
& Invitational and facilitative: Warm, respectful, and encourages reflection. 
& Slightly directive: Helpful but somewhat prescriptive or expert-centered. 
& Dominant, flat, or dismissive: Authoritative, discouraging, or patronizing. \\
\midrule

\textbf{Reflective Prompting} & Does the expert encourage the instructor to reflect or reason?
& Strong probing: Encourages deep reasoning or exploration. 
& Mildly reflective: Prompts some reflection but not in-depth. 
& None or purely directive: No reflection encouraged, just advice or direction. \\
\midrule

\textbf{Appropriateness of Validation} & Does the expert appropriately acknowledge and validate the user’s input without overusing generic praise?
& Balanced and situational: Validation feels genuine and meaningful. 
& Mildly generic or frequent: Some feedback feels formulaic or excessive. 
& Excessive or inauthentic: Overused or insincere validation. \\
\bottomrule
\end{longtable}

\clearpage

\end{document}